\documentclass[10pt,twocolumn,letterpaper]{article}

\usepackage{wacv}              

\usepackage{graphicx}
\usepackage{amsmath}
\usepackage{amssymb}
\usepackage{booktabs}

\usepackage[pagebackref,breaklinks,colorlinks]{hyperref}

\usepackage[capitalize]{cleveref}
\crefname{section}{Sec.}{Secs.}
\Crefname{section}{Section}{Sections}
\Crefname{table}{Table}{Tables}
\crefname{table}{Tab.}{Tabs.}

\def\wacvPaperID{1009} 
\def\confName{WACV}
\def\confYear{2024}

\begin{document}

\title{Efficient Explainable Face Verification based on Similarity Score Argument Backpropagation}

\author{Marco Huber$^{1,2}$, Anh Thi Luu$^{1}$, Philipp Terhörst$^{3}$, Naser Damer$^{1,2}$\\
$^{1}$ Fraunhofer Institute for Computer Graphics Research IGD, Darmstadt, Germany\\
$^{2}$ Department of Computer Science, TU Darmstadt,
Darmstadt, Germany\\
$^{3}$ Paderborn University, Paderborn, Germany\\
Email: marco.huber@igd.fraunhofer.de
}

\maketitle

\begin{abstract}
\vspace{-4mm}
Explainable Face Recognition is gaining growing attention as the use of the technology is gaining ground in security-critical applications.
Understanding why two face images are matched or not matched by a given face recognition system is important to operators, users, and developers to increase trust, accountability, develop better systems, and highlight unfair behavior. 
In this work, we propose a similarity score argument backpropagation (xSSAB) approach that supports or opposes the face-matching decision to visualize spatial maps that indicate similar and dissimilar areas as interpreted by the underlying FR model. Furthermore, we present Patch-LFW, a new explainable face verification benchmark that enables along with a novel evaluation protocol, the first quantitative evaluation of the validity of similarity and dissimilarity maps in explainable face recognition approaches. 
We compare our efficient approach to state-of-the-art approaches demonstrating a superior trade-off between efficiency and performance. The code as well as the proposed Patch-LFW is publicly available at: \url{https://github.com/marcohuber/xSSAB}.
\end{abstract}

\vspace{-6mm}
\section{Introduction} 
\vspace{-2mm}
Automated face recognition (FR) has become an increasingly important part of our lives. It can be used to unlock a smartphone, cross borders at automated border checkpoints, or pay with a face. This is due, among other things, to the ease of use and high accuracy of modern FR systems. In recent years, the high accuracy of biometric systems has been driven primarily by larger databases \cite{DBLP:conf/iccvw/AnZGXZFWQZZF21}, innovative solutions \cite{elasticface, magface}, and advances in deep learning \cite{resnet, DBLP:conf/iclr/DosovitskiyB0WZ21,EFAR,DBLP:journals/access/BoutrosSKDKK22}.  

However, the methods based on deep learning have the disadvantage of being difficult to understand because they include millions of parameters and are highly complex models \cite{pedro}.
For various reasons, biometric systems require that they are more understandable to humans. Understanding increases trust, can highlight unfair or unequal behavior toward different subgroups, or help develop better systems \cite{DBLP:journals/pr/0001WLLSSK21, DBLP:journals/access/AdadiB18}.

The reasons or clues why two face images are determined to be a match or non-match by an FR system have received increasing attention recently \cite{xCos, bbox, bmvc, truebbox, xface}. Especially in security-relevant areas, but also for usability, it is interesting to understand why two images are falsely recognized as a match or falsely recognized as a non-match (e.g. why didn't I match my passport at the automatic border control gate?). 

\begin{figure}
    \centering
    \includegraphics[width=\columnwidth]{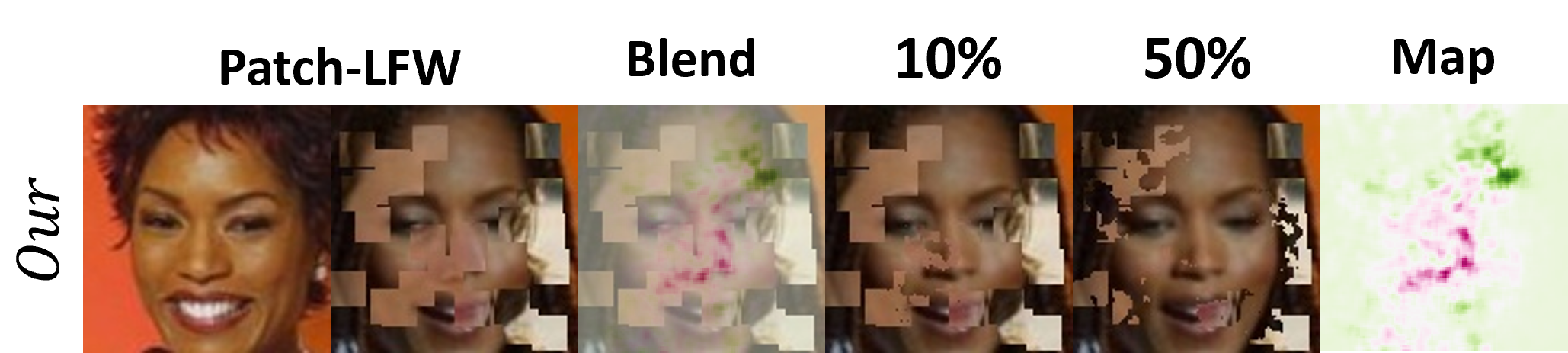}
    \caption{An example visualization of our proposed efficient explainable face verification explanation, xSSAB, on the novel Patch-LFW benchmark and its evaluation protocol. The added patches from an imposter identity on an original genuine pair shifted the model's decision to non-match. During the evaluation, the added imposter clues (patches) are replaced by the original pixels based on the identified dissimilar area (pink) of our approach.} 
    \label{fig:catcher}
    \vspace{-5mm}
\end{figure}

In the field of biometrics, explainability and interpretability of biometric decisions and systems have been identified as outstanding problems \cite{trust}. In the area of more understandable face matching decisions, Lin et al. \cite{xCos} proposed xCos. Their approach is based on a novel more interpretable cosine metric that provides meaningful explanations. Huber et al. \cite{bmvc} have proposed modeling the uncertainty and confidence of model decisions using stochastic forward passes to gain more insight into the decision process. In comparison, Knoche et al. \cite{xface} proposed a confidence score and a visualization approach based on systematic occlusions. {\color{black} A current trend is to follow a black-box approach \cite{xface,truebbox, bbox}, where only the input is changed and the changes in the output are observed, without a deeper understanding of the inner workings of the model. While these approaches provide meaningful explanations, they require plenty of forward calculations which are time-consuming. Different approaches are white-box explanation approaches \cite{xCos}, which require access to the model, but can be more efficient. Our proposed approach follows the white-box approach to provide fast explanations without additional training.} 

One remaining challenge in the field of explainable FR is having an evaluation protocol to assess the validity of the explanations. Often, they are evaluated only on the basis of visualizations on a subjective basis without a quantitative evaluation \cite{xface, xCos, bbox, truebbox}, which makes it difficult to compare different approaches objectively.

In this work, we 1) propose a novel, efficient, and training-free {\color{black} white-box} approach based on similarity score backpropagation to indicate areas in a pair of face images to explain, which parts of the image are interpreted as similar or dissimilar for an FR model, with an example shown in Figure \ref{fig:catcher}, 2) propose an objective evaluation dataset and an evaluation protocol that allows comparing the performance of different explainable face matching approaches based on visualization. 

{\color{black} Our contributions are:
\begin{itemize}
\vspace{-2mm}
    \item 1) A training-free explainable FR approach that is competitive with state-of-the-art (SOTA) on several FR models while being time-efficient
    \vspace{-3mm}
    \item 2) The first explainable FR benchmark, Patch-LFW, including an evaluation protocol that allows to compare explainable face matching approaches in an objective manner.
\end{itemize}}
\vspace{-5mm}
\section{Related Work} 
\label{related}
\vspace{-2mm}
In recent years, the performance of FR systems has been improving and has already surpassed the performance of humans \cite{human}. This improvement is based on larger data sets, larger models, and new loss functions, such as ArcFace loss \cite{arcface}, CurricularFace loss \cite{curricularface}, MagFace loss \cite{magface} or ElasticFace loss \cite{elasticface}. With the increase in verification performance, the models became less understandable due to their complexity and size, which raises questions about the inner workings and causes of decisions.

The first direction of research that focused on making the face matching process more understandable were works related to uncertainty mapping. The idea for uncertainty mapping comes from thinking that faces may be ambiguous or lack identity information. Moreover, the problem of model and data uncertainty is of general interest in the deep learning community \cite{DBLP:conf/nips/KendallG17, DBLP:conf/icml/GalG16}. Shi and Jain \cite{pfe} proposed probabilistic face embeddings, where each face image is represented as a Gaussian distribution with the feature as the mean and the uncertainty of the features as the variance. This general idea has then been adopted in several ways \cite{dataunc, fastunc}, including propagating the uncertainty present in the process of face matching to the comparison itself \cite{bmvc, DBLP:conf/iwbf/HuberTKKD23}. Others tried to explain the performance variations in FR over different demographic groups \cite{DBLP:conf/icb/FuD22}, or assign the different in explainability performs \cite{10289865} as well as the different levels of face image quality \cite{DBLP:journals/tbbis/TerhorstHDKRK23, DBLP:conf/wacv/FuD22}.

The second research direction towards explainable FR followed the trend in computer vision to visualize important areas using saliency or heatmaps \cite{gradcam, scorecam}, there are also works that visualize the crucial areas in the decision-making of FR. Applying methods known from computer vision like GradCAM \cite{gradcam} or Score-CAM \cite{scorecam} are not natively applicable for FR models in an optimal manner, since they are designed for classification problems and not for a process that consists of feature (embedding) extraction followed by embedding comparison (matching), which is typically applied in SOTA FR systems.

To visualize the important areas for an FR model, Lin et al. \cite{xCos} proposed a novel similarity metric named xCos based on a learnable module to provide meaningful explanations. Their approach can be applied to most of the verification models, however, the module has to be trained, which reduces its adaptability, if the underlying model is often changed. 
{\color{black} Knoche et al. \cite{xface}, Mery \cite{truebbox} and Mery and Morris \cite{bbox} proposed to explain black-box FR models, following a justification explainability approach \cite{DBLP:conf/cvpr/ParkHARSDR18} designed to visually communicate the decisions' evidence instead of an introspective approach reflecting the inner workings visually \cite{DBLP:conf/cvpr/ParkHARSDR18}. Their approaches are designed to work without access to the FR model and are based on perturbing or altering the input face images and investigating changes in the output.} While this produces interpretable saliency maps, the calculation is time-consuming. This poses a problem in practical applications, for example, if the user has to wait a long period of time for a reaction from the system, or if the data has to be transferred to a more powerful system in order to be processed there. 


All the saliency map-producing approaches mentioned above did not evaluate their approaches in a quantified way, but rather limited their evaluation to visualizations, making an assignment and comparison in terms of the correctness and quality of the proposed approaches hard. They also did not present or use a well-defined benchmark specifically designed to demonstrate their results in a comparable manner. 

\vspace{-2mm}
\section{Methodology}
\label{method}
\vspace{-2mm}

\begin{figure*}
    \centering
    \includegraphics[width=0.6\textwidth]{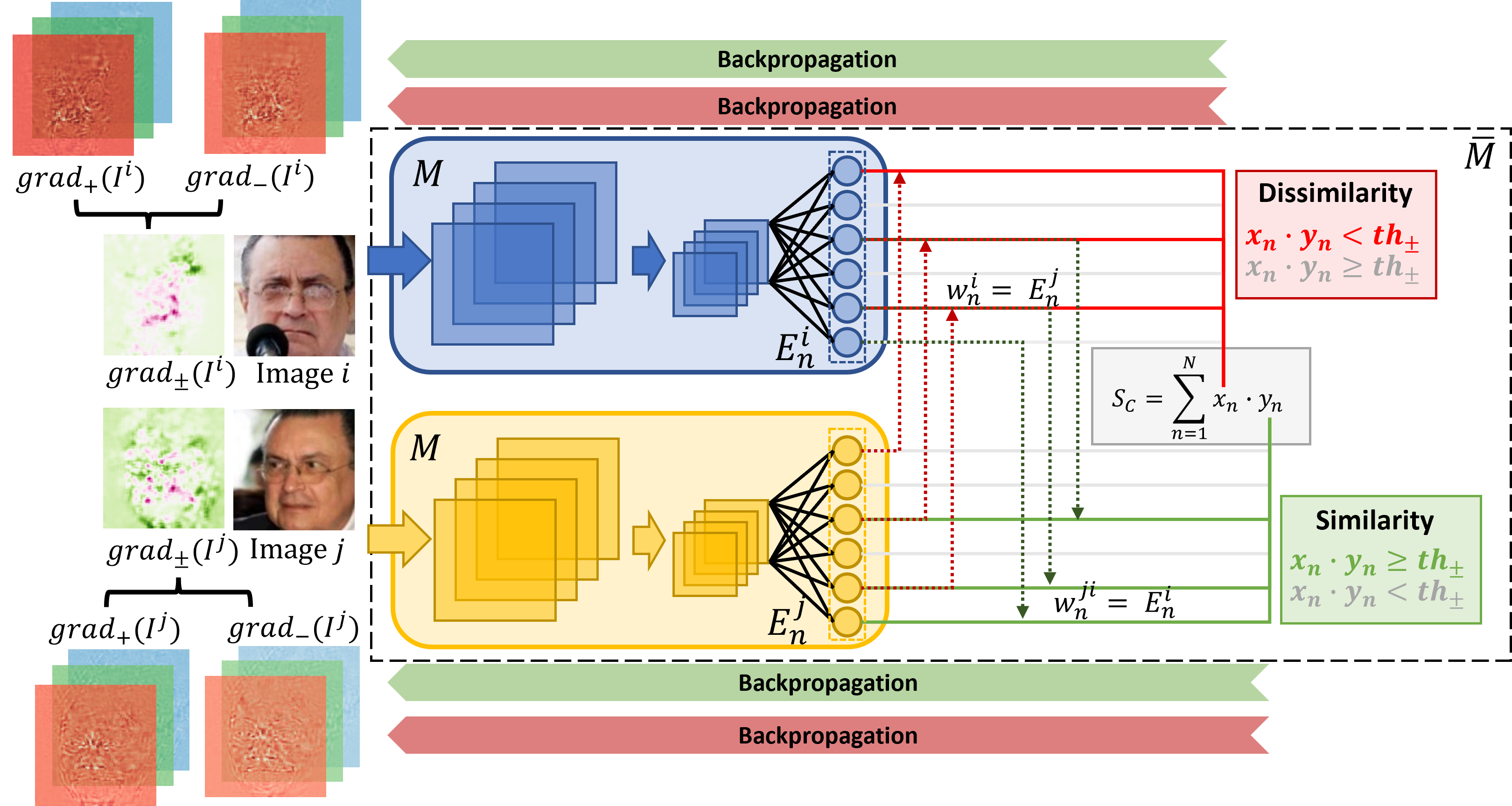}
    \caption{Overview of the proposed approach: In a siamese fashion, both face images are processed by a FR system $M$ that is extended with an additional cosine similarity layer, {\color{black} which simply calculates the cosine similarity between the face embeddings.}. Based on the system decision threshold and the contribution of the different features to the similarity score, gradients are backpropagated to obtain maps that highlight similarity and dissimilarity. These maps are then fused in a final step to get a single map, highlighting both.} 
    \label{fig:overview}
    \vspace{-4.5mm}
\end{figure*}

In this section, we present and rationalize our proposed xSSAB approach to explain face verification decisions efficiently. Understanding which parts of an image pair lead to a matching or non-matching decision is crucial to make FR more transparent and explainable. We propose to backpropagate the similarity score of face image pairs through a Siamese FR setup to efficiently indicate which parts of the image pair contribute to a match or non-match decision. By investigating the positive and negative impact of feature dimensions on the similarity-based comparison score, gradients based on these impacts can be backpropagated to highlight important pixels. An illustration of our methodology is shown in Figure \ref{fig:overview}.

In the first step, we propagate both images through a siamese network of FR model $M$ to gain feature embeddings $E^{i}, E^{j}$ of the face images $I^{i}, I^{j}$ as demonstrated in Figure \ref{fig:overview}. We extend the FR model $M$ by adding a cosine similarity layer, which calculates and outputs the cosine similarity $S_{C}$ of both input face images to include the matching process in the network architecture. To simplify the calculation we also internally normalize the {\color{black}embedding vectors, $E^i$ and $E^j$,} before calculating the cosine similarity which simplifies the cosine similarity formula:
{\color{black}
\begin{equation}
\footnotesize
    S_{C}(E^i,E^j) = \frac{E^i \cdot E^j}{\lVert E^i \lVert \lVert E^j \lVert},
\end{equation}
to
\begin{equation}
\footnotesize
    S_{C}(E^i,E^j) = \sum_{n=1}^{N} E^{i}_{n} \cdot E^{j}_{n},
    \label{simpcos}
\end{equation}
as $\lVert E^i \lVert \lVert E^j \lVert= 1$. $N$ denotes the number of dimensions of the embeddings, $E^i$ and $E^j$. $E^{i}_{n}$ and $E^{j}_{n}$ denotes the feature value at position $n$ of $E^i$ or $E^j$, respectively.} Our extended FR model $\overline{M}$ can therefore be defined as:
\begin{equation}
\footnotesize
    \overline{M}_{i,j} = \sum M(I^{i}) \cdot M(I^{j}),
\end{equation}
where $M(I^{i})$ is similar to the embedding $E^{i}$ and $M(I^{j})$ to the embedding $E^{j}$:
\begin{equation}
\footnotesize
    \overline{M}_{i,j} = \sum_{n=1}^{N} E^{i}_{n} \cdot E^{j}_{n}.
\end{equation}
Since $\overline{M}_{i,j}$ includes weights dependent on the provided input images to calculate the cosine similarity ($\overline{M}_{i,j}$ output) of the provided input images, $I^{i}$ and $I^{j}$, it is dependent on the input.

Given the simplified cosine similarity function (Equation \ref{simpcos}), we can observe that, expectedly, the cosine similarity will increase if the feature values $x_{n}$ and $y_{n}$ share the same direction and, otherwise, decrease the score. Since the final decision of match or non-match is not just dependent on the similarity score but also includes the system-dependent decision threshold $th_{d}$, we include this in the calculation of our approach. With the assumption, that each feature in the different feature dimension ideally at least contributes equally to the matching decision and therefore to the similarity score, we define an argument threshold $th_{\pm}$. This $th_{\pm}$ is set to:
\begin{equation}
\footnotesize
    th_{\pm} = \frac{th_{d}}{n},
\end{equation}
where $n$ refers to the number of dimensions in the feature space. We then define a feature argument $a_{n}$ for each feature dimension $n$ as:
{\color{black}
\begin{equation}
\footnotesize
    a_{n} = E^{i}_{n} \cdot E^{j}_{n}.
\end{equation}}
The interpretation is then that the feature argument $a_{n}$ in the feature dimension $n$ with a positive impact on the score ($a_{n} \geq  th_{\pm} $) provides a \textit{positive argument} and with a negative impact on the score ($a_{n} \le  th_{\pm} $) as provides a \textit{negative argument}. 
The intuition is, that the argument threshold $th_{\pm}$ defines the minimum strength of an argument $a_{n}$ to be considered a positive argument ($a_{n} \in a_{+}$) depending on the system-dependent threshold $th_{d}$. This is the case if the argument is at least as strong as it has to be if all other arguments equally contribute to a matching decision.  In the other case, it is considered a negative argument ($a_{n} \in a_{-}$). 

Starting from the calculated positive argument set $a_{+}$ and negative argument set $a_{-}$, we calculate the gradient based on the arguments, either only for the set of positive arguments ($a_{+}$) or for the set of negative arguments ($a_{-}$). We calculate the gradients backward through the FR model $\overline{M}_{i,j}$ to obtain the pixels that have the most influence \cite{backprop} on the positive or negative arguments given both images $I_{i}$ and $I_{j}$. To limit the calculation to only the impact on the positive or negative arguments, we manipulate the weight $w^i$ of the cosine layer of $\overline{M}_{i,j}$ (which is originally equal to $E^j$, since it computes $S_C$), so only the weights are included in the forward and backpropagation process that are either positive or negative:
\begin{equation}
\footnotesize
    w^i_n = \begin{cases} E^j_n & \text{if } a_n \in a_{+} \\
    0, & \text{otherwise}
    \end{cases} ,
\end{equation}
if we want to obtain the similarity map of image $I^i$ given image $I^j$ and 
\begin{equation}
\footnotesize
    w^i_n = \begin{cases} E^j_n & \text{if } a_n \in a_{-} \\
    0, & \text{otherwise}
    \end{cases} ,
\end{equation}
if we want to obtain the dissimilarity map of image $I^i$ given image $I^j$. We define the models with the adjusted weights $w^i_n$ as $\overline{M}^{+}_{i,j}$ in the first case and $\overline{M}^{-}_{i,j}$ in the second case. Since we manipulate the weights depending on the given comparison image, we get two different models and maps for the images $I^{i}$ and $I^{j}$.

To calculate the gradient based on the positive ($grad_+$) and negative arguments ($grad_-$), we compute:
\begin{equation}
\footnotesize
    grad_+(I^{i}) = \frac{\partial \overline{M}^{+}_{i,j}}{\partial I^{i}},
\end{equation}
for the similarity map and
\begin{equation}
\footnotesize
    grad_{-}(I^{i}) = \frac{\partial\overline{M}^{-}_{i,j}}{\partial I^{i}},
\end{equation}
for the dissimilarity map. This allows us to calculate gradient-based maps only based on the features that have a positive or negative impact regarding a matching or non-matching decision based on the system's decision threshold.

To optimize the visualization further, we take the mean of the gradients of the three color channels ($c$). To be independent of the sign of the gradients, we also take the absolute value:
\begin{equation}
\footnotesize
    \overline{grad}_{+}(I^{i}) = \frac{1}{3} \sum_{c=1}^{3} \lvert grad_+,c(I^{i})\rvert,
\end{equation}
for the positive explanation map $\overline{grad}_{+}$. The negative (non-matching) map, $\overline{grad}_{-}$, is calculated in the same manner, only based on $grad_-,c(I^{i})$.
Our approach provides us with two explanation maps, showing which pixels lead to a positive or negative argument regarding the final matching decision of the FR system depending on the given decision threshold $th_{d}$ and the investigated face images $I^{i}$ and $I^{j}$. To get a single explanation map, $\overline{grad}_{\pm}$, explaining both, similarities and dissimilarities, we combine the two calculated explanations maps naturally by subtracting the negative explanation map from the positive explanation map:
\begin{equation}
\footnotesize
    \overline{grad}_{\pm} = \overline{grad}_{+} - \overline{grad}_{-}.
\end{equation}
To have a less fragmented visualization, we finally apply a Gaussian blur filter with $5 \times 5$ filter and with $\sigma = 5$, similar to Knoche et al. \cite{xface}.

The proposed xSSAB approach, therefore, generates a single face verification explanation map based on the pixels' influence on the final matching or non-matching decision based on the system-dependent decision threshold. To achieve this, gradients based on the similarity score arguments are utilized. Calculating the explanation maps rather based on the internal behavior than just interpreting the model in a black-box fashion and altering the input, allows more efficient transparency of the models' behavior.

\vspace{-2.5mm}
\section{Experimental Setup} 
\label{experiments}
\vspace{-1.5mm}
\subsection{Face Recognition Models}
\vspace{-1.5mm}
To show the validity and the generalizability of our approach, we utilize four different SOTA FR models in our experiments. All models share the same ResNet-100 \cite{resnet} architecture and have been trained with the corresponding loss functions. All the models were trained on the MS1M-V2 \cite{arcface} dataset. The used models are: ArcFace \cite{arcface}, ElasticFace-Cos \cite{elasticface}, ElasticFace-Arc \cite{elasticface}, and CurricularFace \cite{curricularface}, and they are all used as pre-trained models provided in their respective official repositories. We utilize these models to show that our approach can be applied to a wide range of diverse FR models without training or fine-tuning {\color{black} and the models are competitive to other SOTA models such as MagFace \cite{DBLP:conf/cvpr/MengZH021} or AdaFace \cite{DBLP:conf/cvpr/Kim0L22}}.
\vspace{-1.5mm}
\subsection{Evaluation Benchmark: Patch-LFW}
\begin{figure}
     \centering
     \begin{subfigure}[b]{0.2\textwidth}
         \centering
         \includegraphics[width=\textwidth]{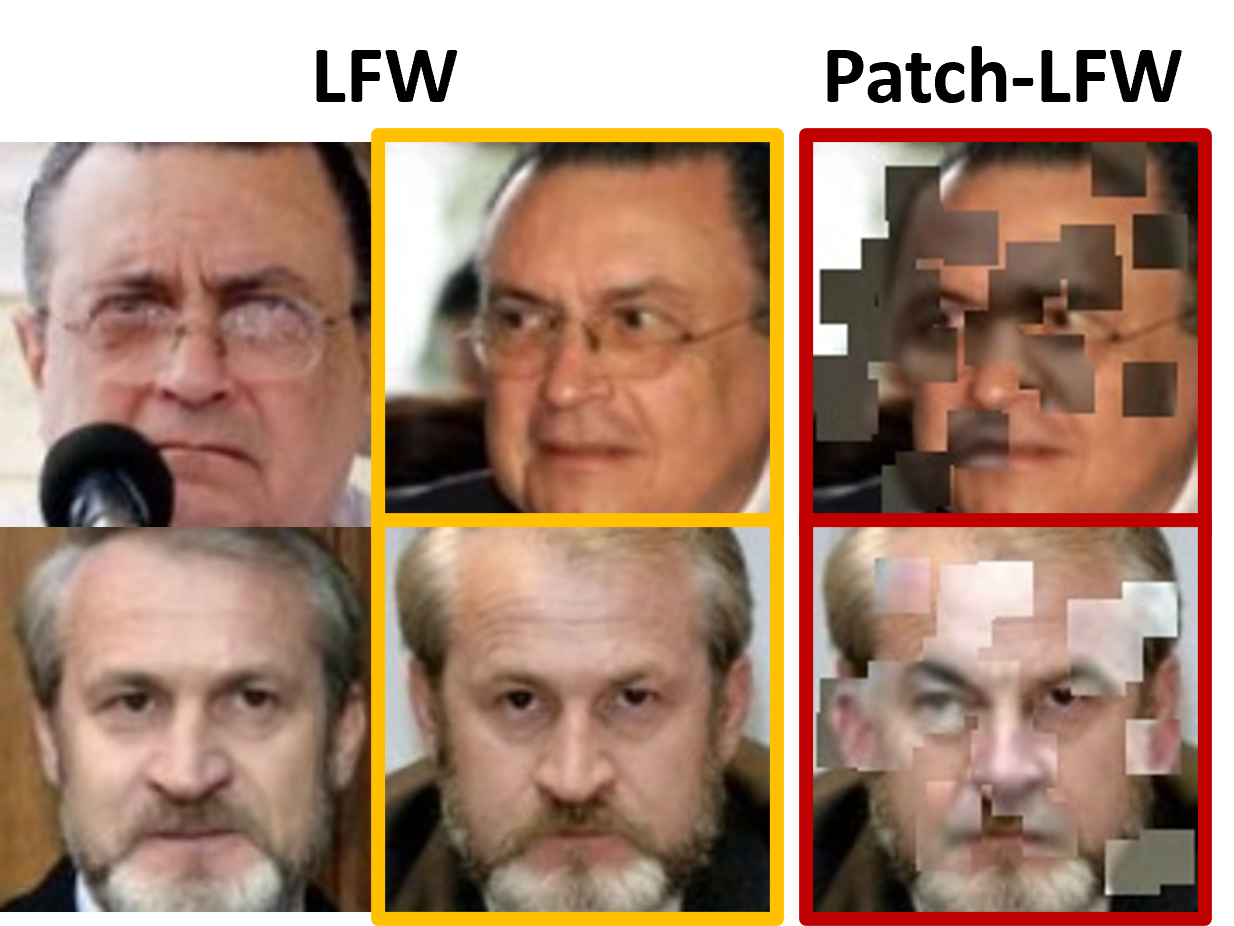}
         \caption{Genuine Pairs}
         \label{fig:gen_pair}
     \end{subfigure}
     \begin{subfigure}[b]{0.2\textwidth}
         \centering
         \includegraphics[width=\textwidth]{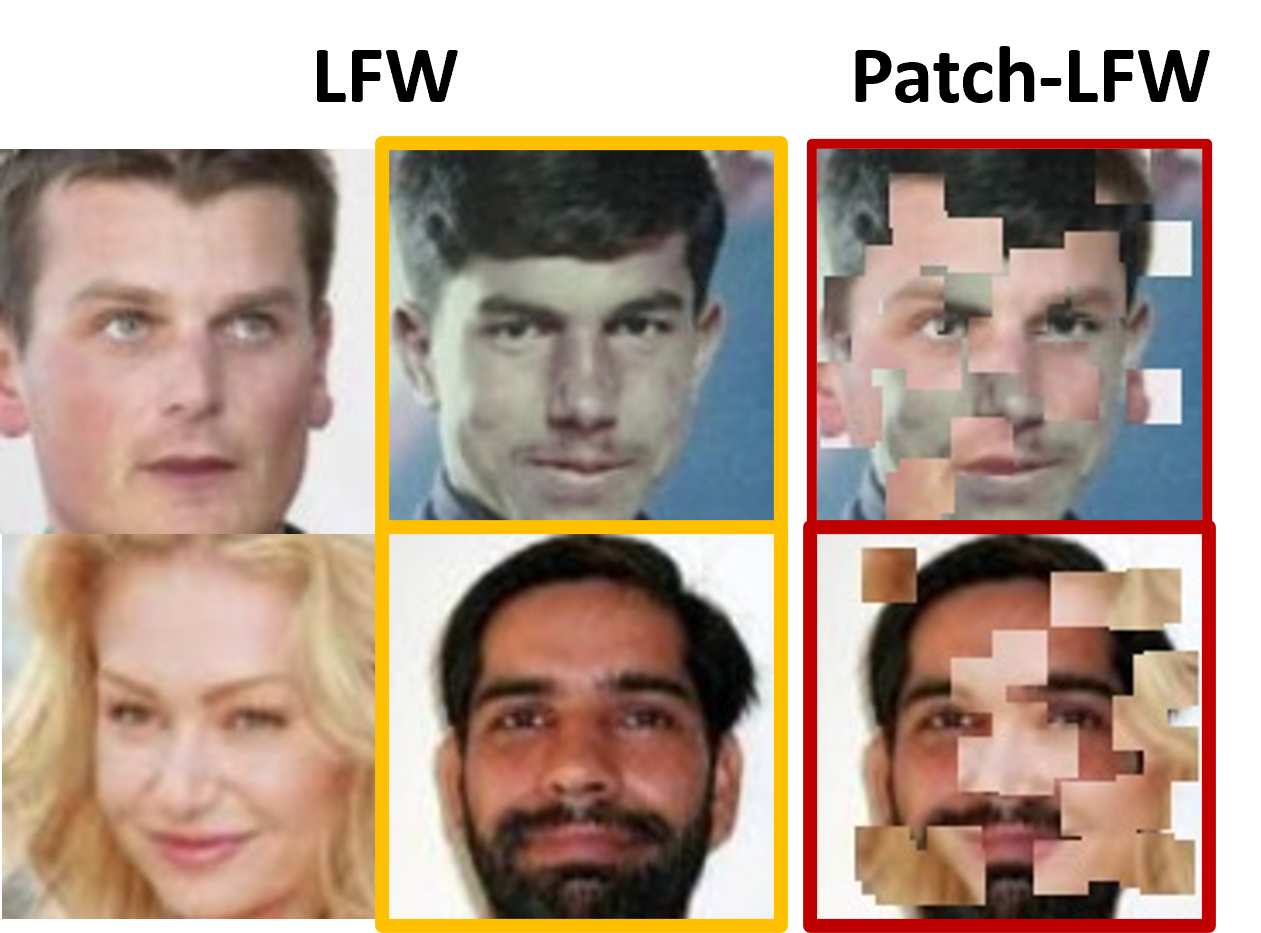}
         \caption{Imposter Pairs}
         \label{fig:imp_pair}
     \end{subfigure}
        \caption{Example pairs from Patch-LFW: the images with the yellow outline in LFW are in Patch-LFW replaced by the images with the red outline. By adding information from a random imposter identity (genuine case) or from the same image (imposter case) the amount of false matches (FM) and false non-matches (FNM) increases.} 
        \label{fig:pair} 
        \vspace{-4mm}
\end{figure}

\vspace{-1.5mm}
For the evaluation of the proposed explainable face matching approach and the comparison with other methods, we build a new benchmark dataset, \textit{Patch-LFW}. Patch-LFW is based on the Labeled Faces in the Wild (LFW) \cite{lfw} dataset. Since in most cases it is more interesting, from a practical perspective, to know why the system made a wrong decision than to understand why it made the right decision, we artificially increase the errors that the system makes. 
{\color{black} The reason we chose LFW as the starting point for our Patch-LFW is the simplicity of the dataset. To measure how well we explain wrong decision reasons, we need to minimize unknown reasons for wrong decisions and only have the ones we control, thus we need a benchmark that SOTA FR makes as minimum as possible wrong decisions. SOTA systems solve the verification problem in the LFW dataset almost perfectly, which minimizes other influences apart from the patches we add, which is what we need. Choosing a more challenging dataset as the baseline would make it harder to distinguish if the approaches can identify the added deterioration (patches) or inherent "clues" such as bias \cite{9534882} or occlusions \cite{DBLP:journals/iet-bmt/0002VS21,DBLP:journals/access/NetoPBDSC22}}. To synthetically increase the amount of "false non-matches", in each of the genuine pairs, patches from a random image of a different identity have been added to the reference image. To synthetically increase the amount of "false matches", in each of the imposter pairs, patches from the same image have been added to the reference image. In total, we randomly added 27 patches of size $16 \times 16$  pixels which may overlap per image. Before adding the patches, the images of LFW \cite{lfw} are preprocessed following the procedure of Deng et al. \cite{arcface}. Examples of the newly created images and the original images from LFW \cite{lfw} are shown in Figure \ref{fig:pair}. Given the new Patch-LFW, the "false match" rate as well as the "false non-match" rate at a fixed threshold determined at the Equal Error rate (EER) \cite{iso_metric} on using the whole LFW dataset \cite{lfw} increased drastically as shown in Table \ref{tab:patch}, which is our goal. 

\subsection{Evaluation Protocol: Decision-based Patch Replacement}
\label{evalprot}
\begin{figure*}
     \centering
     \begin{subfigure}[b]{0.24\textwidth}
         \centering
         \includegraphics[width=\textwidth]{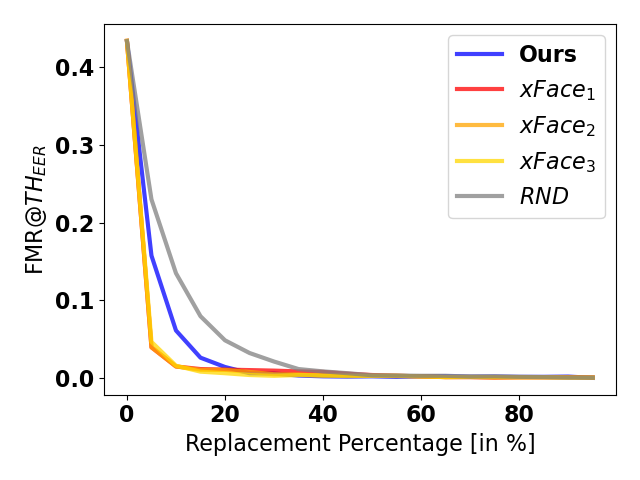}
         \caption{ElasticFace-Arc - FMR}
         \label{fig:elasticarc-fmr}
     \end{subfigure}
     \begin{subfigure}[b]{0.24\textwidth}
         \centering
         \includegraphics[width=\textwidth]{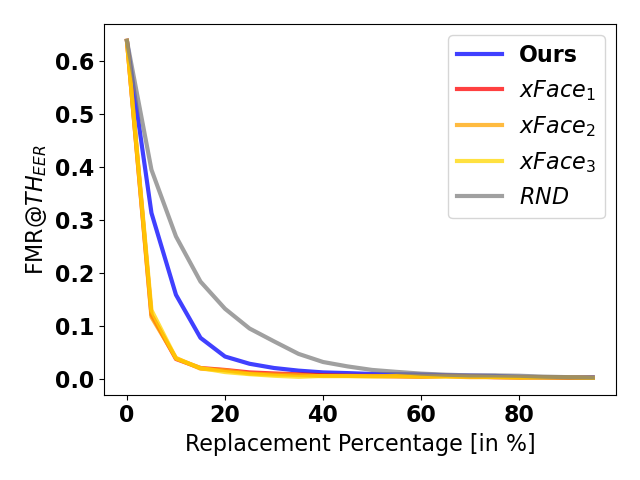}
         \caption{ElasticFace-Cos - FMR}
         \label{fig:elasticcos-fmr}
     \end{subfigure}
          \begin{subfigure}[b]{0.24\textwidth}
         \centering
         \includegraphics[width=\textwidth]{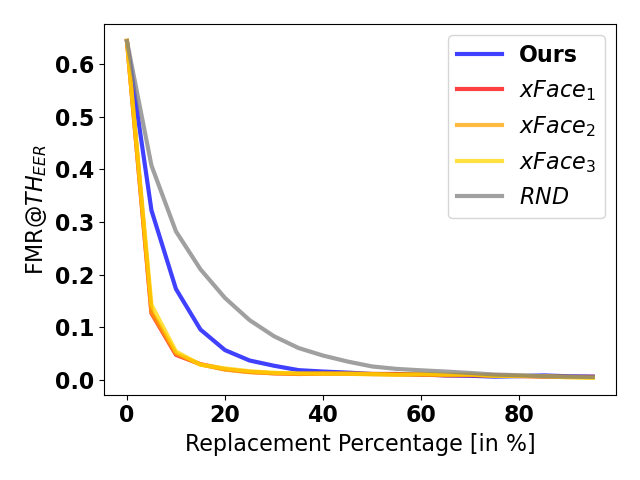}
         \caption{CurricularFace - FMR}
         \label{fig:curr-fmr}
     \end{subfigure}
     \begin{subfigure}[b]{0.24\textwidth}
         \centering
         \includegraphics[width=\textwidth]{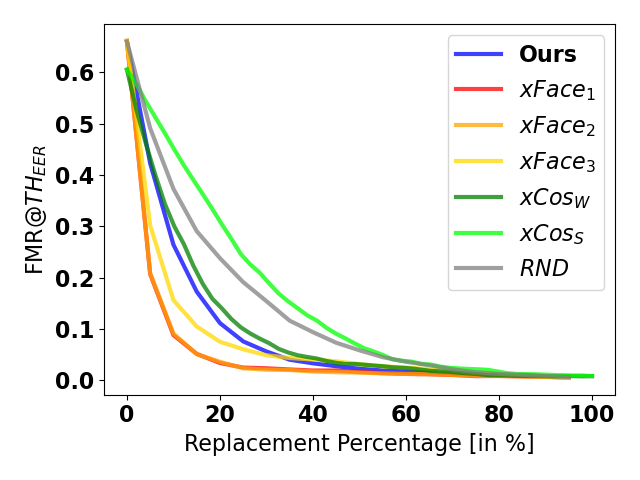}
         \caption{ArcFace - FMR}
         \label{fig:arc-fmr}
     \end{subfigure}
          \begin{subfigure}[b]{0.24\textwidth}
         \centering
         \includegraphics[width=\textwidth]{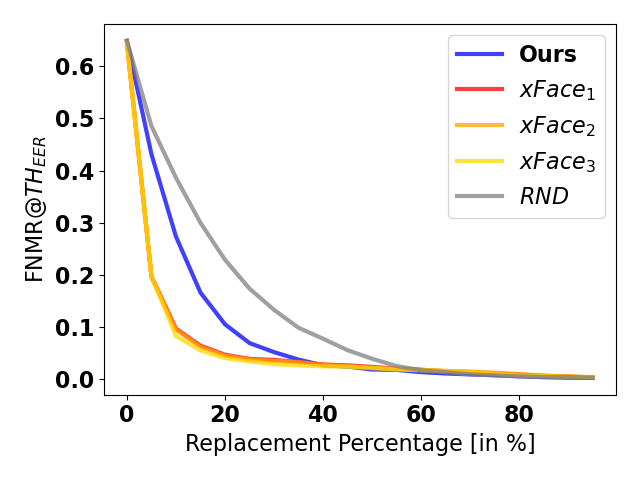}
         \caption{ElasticFace-Arc - FNMR}
         \label{fig:elasticarc-fnmr}
     \end{subfigure}
     \begin{subfigure}[b]{0.24\textwidth}
         \centering
         \includegraphics[width=\textwidth]{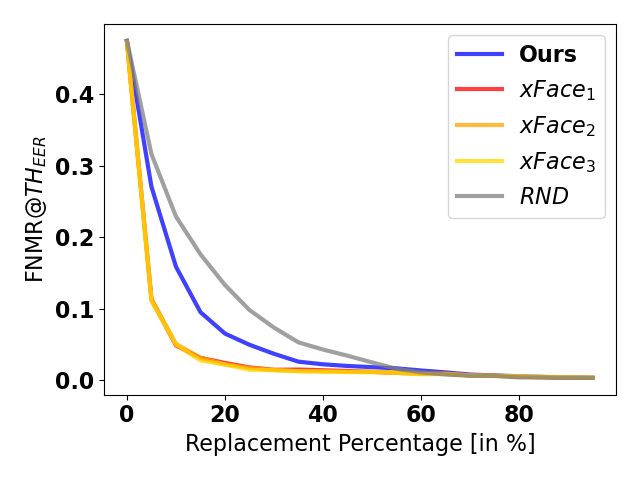}
         \caption{ElasticFace-Cos - FNMR}
         \label{fig:elasticcos-fnmr}
     \end{subfigure}
          \begin{subfigure}[b]{0.24\textwidth}
         \centering
         \includegraphics[width=\textwidth]{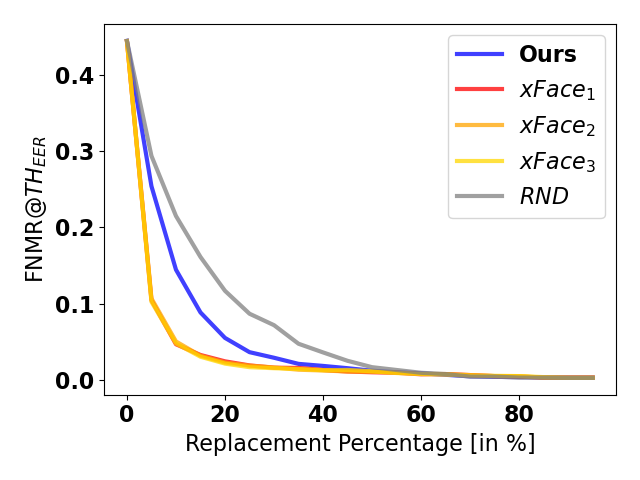}
         \caption{CurricularFace - FNMR}
         \label{fig:curr-fnmr}
     \end{subfigure}
     \begin{subfigure}[b]{0.24\textwidth}
         \centering
         \includegraphics[width=\textwidth]{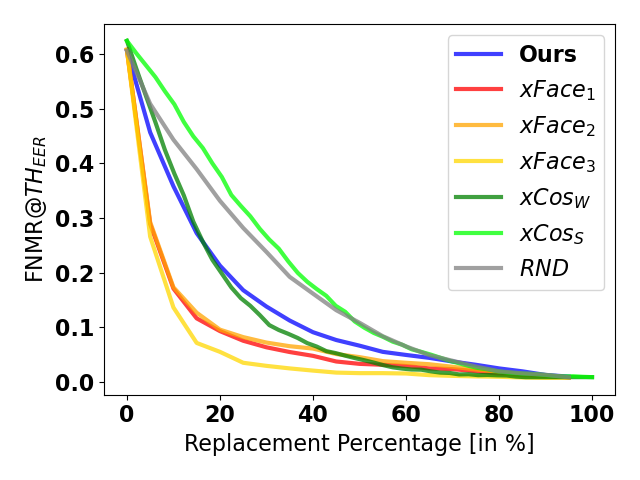}
         \caption{ArcFace - FNMR}
         \label{fig:arc-fnmr}
     \end{subfigure}
     \vspace{-2mm}
        \caption{Decision-based Patch Replacement (DPR) curves on Patch-LFW: The error in terms of FMR and FNMR at the fixed threshold $th_{EER}$ determined on LFW, based on the percentage of replaced pixels. Our approach outperforms the random baseline ($RND$) on all errors and on all models. Our approach shows to be applicable to a wide range of diverse FR models. Even though the $xFace$ methods show slightly better performance, they are not as efficient as discussed in Section \ref{latency}. On the Arcface model, our approach outperforms the pre-trained $xCos_{S}$ approach and is comparable to pre-trained $xCos_{W}$, while being training-free. We did not evaluate the $xCos$-based approaches on other models, since the pre-trained module did not provide competitive results on the other models, proofing that it is necessary to train the $xCos$ module before. The AUC values are provided in Table \ref{auc}.} 
        \label{fig:dpr}
        \vspace{-3mm}
\end{figure*}

To quantitatively determine the quality of explanation maps, we evaluate on our newly proposed Patch-LFW with a novel evaluation method (Decision-based Patch Replacement (DPR) curve). The DPR curves are inspired by the Insertion and Deletion curves \cite{DBLP:conf/bmvc/PetsiukDS18}. First, the similarity maps for the original image pairs in the Patch-LFW dataset are determined. Based on these maps and the decision threshold $th_{EER}$ determined on the original LFW without patches {\color{black} for each FR model}, we proceed as follows:
\begin{itemize}
    \item 1)  If the patched image pair is considered a match based on $th_{EER}$, the most similar pixels are adjusted based on the explanation map. If the patched image pair is considered a non-match, the least similar pixels are adjusted. During the adjustment, the identified pixels are replaced by the original pixels from LFW.
    \item 2) After adjusting a fixed amount of pixels (in our case 5\%), the FR systems are utilized to evaluate the new performance in terms of FMR and FNMR. If the explainability approach is of high quality, it should be able to detect the areas (added patches) that lead to "false matches" and "false non-matches" and reduce the errors when more and more pixels are replaced. If a pixel is detected to be replaced that is original and not part of an added patch, it is left unchanged. This step is repeated until all pixels have been replaced.
\end{itemize}
To determine the quality, we now look at the drop in error rates over the proportion of pixels replaced. If the curve is lower, it indicates a better performance of the explanation map as the error rate is lower. This proves that the explanation map detected the artificially inserted patches and replaced them with the original pixels, which removes the clues that led to the artificially created error. Such plots are presented in Figure \ref{fig:dpr} and will be discussed in Section \ref{results}.

\vspace{-1mm}
\subsection{Explainable Face Verification Methods}
\vspace{-1mm}
\begin{figure}
    \centering
    \includegraphics[width=0.45\columnwidth]{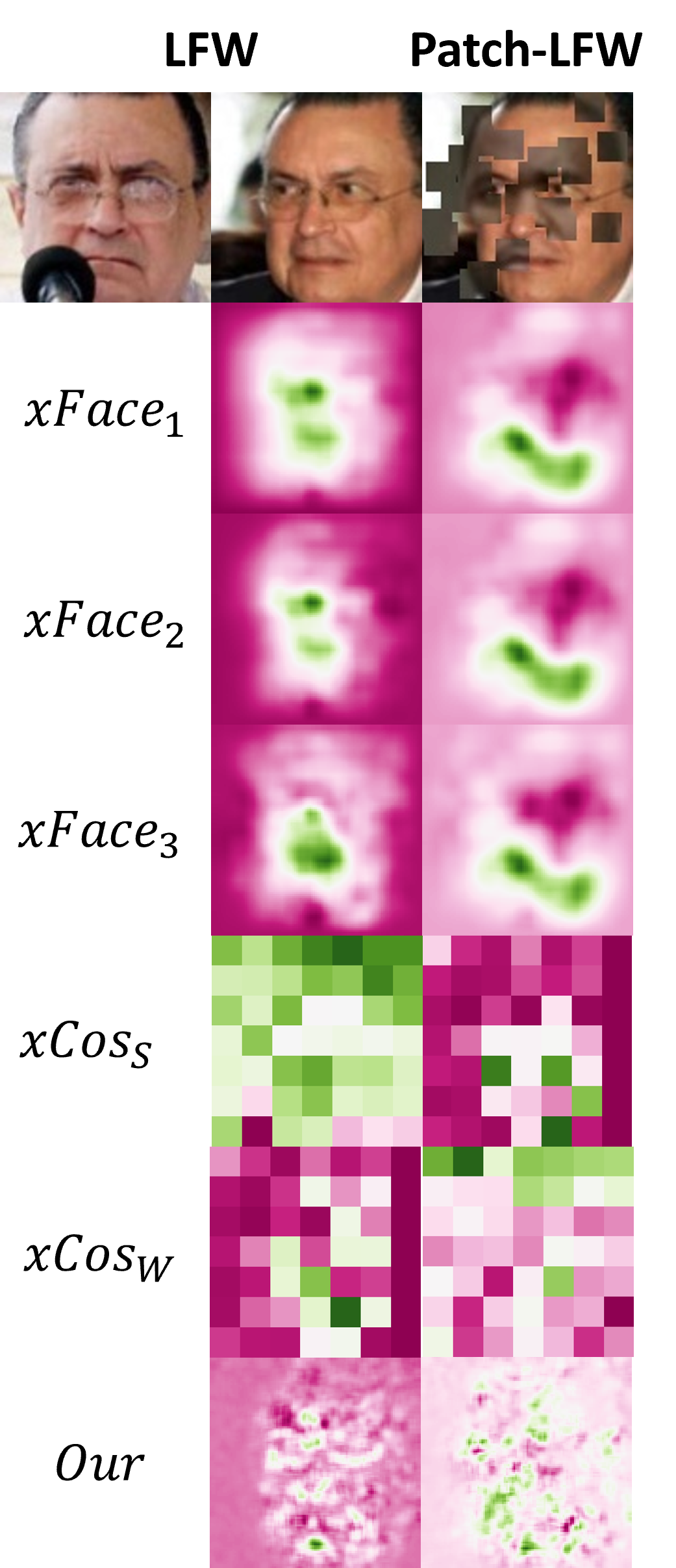}
    \caption{Comparison of different approaches: explanation maps visualized for different approaches. Pink areas indicate dissimilarity, and green areas indicate similarity. The values are not normalized. The design of the maps is very different. While the $xFace_{x}$ approaches tend to highlight contiguous areas, the highlighting in $xCos_{x}$ is patch-wise. Our approach, on the other hand, is more fine-grained. {\color{black}More examples are provided in the supplementary material.}}
    \label{fig:comparison} 
    \vspace{-4mm}
\end{figure}

To evaluate our proposed efficient explainable face verification approach based on similarity score argument backpropagation (xSSAB), we compare our approach in terms of quality and latency with two SOTA solutions: xCos \cite{xCos} and xFace \cite{xface}. 

xCos \cite{xCos} modifies the backbone of the model with a $1\times1$ convolution to preserve the position information of each feature. With an additional attention mechanism, two outputs are retrieved: an attention weight map $W$ and a patched cosine map $S$. Since the authors only provided a pre-trained version for an ArcFace model \cite{arcface}, we stick to the comparison with this model. We experimentally tested to use the proposed pre-trained cosine module and metric on the other models as well, but the equal error rate on ElasticFace-Arc \cite{elasticface} was 0.2533, which is not competitive when compared to the performances presented later in Section \ref{results}. We investigated the performance by inserting pixels either based on the weight map ($xCos_{W}$) or based on the patches cosine map ($xCos_{S}$). Because both of the maps provide equally-sized patches that share the same value, we do not iteratively replace 5\% of the pixels but replace them patch-wise. An example of both maps is shown in Figure \ref{fig:comparison}.

xFace \cite{xface} follows the black-box approach and considers the FR models as pure input-output functions. The core principle of their proposed approach is based on occluding parts of the image and investigating the deviation of the occluded and non-occluded images. In total, they proposed three variations of their approach, which we refer to as $xFace_{1}$, $xFace_{2}$, and $xFace_{3}$. For the parameters of their approach, we follow their proposed values and select stride $s$ = 5 and patch sizes $p \in \{7,14,28\}$. An example of the visualization of the maps can be seen in Figure \ref{fig:comparison}.

We also compare our approach against a random approach ($RND$). The random approach randomly selects pixels to replace and serves as a minimum baseline.

\vspace{-2mm}
\subsection{Evaluation of Latency}
\label{latency}
\vspace{-2mm}
To evaluate the efficiency, we perform the computations on the same machine equipped with an NVIDIA Quadro P4000 GPU and an Intel Core i9-7920X CPU. We use the reference implementations of the respective authors and only consider the time for creating the similarity maps. For the latency determination, we use 200 random images from LFW, in total 100 image pairs. To get as close as possible to a realistic application for an end user, we also evaluate the time for a run and process the pairs in pairs and not batch-wise. We report the total time over the 100 image pairs as well as the average time per processed pair. For an end-user application scenario, the user (whether an individual or operator) would expect a timely response, as being verified is not his main goal but e.g. crossing a border or getting access to a restricted area/device.

\begin{table}[]
\centering
\small
\begin{tabular}{c|l|ll}
           & LFW                     & \multicolumn{2}{c}{Patch-LFW}                      \\ \hline
Model           & \multicolumn{1}{c|}{EER} & \multicolumn{1}{c}{FMR} & \multicolumn{1}{c}{FNMR} \\ \hline
ElasticFace-Arc \cite{elasticface} &  0.003                       & 0.434                       & 0.649                         \\
ElasticFace-Cos \cite{elasticface} & 0.002                        & 0.639                        & 0.475                         \\
CurricularFace \cite{curricularface} & 0.005                        & 0.643                        & 0.444                         \\
ArcFace \cite{arcface} & 0.005                         & 0.660                        & 0.606                        \\ \hline
\end{tabular}
\vspace{-3mm}
\caption{Performance of the utilized models on LFW \cite{lfw} and the proposed benchmark Patch-LFW. The FMR and FNMR at Patch-LFW have been evaluated at the EER threshold {\color{black} $th_{EER}$} determined on LFW. {\color{black}By definition, the FMR and FNMR of LFW are equal to the EER.} The results show that the introduced patches increased the FMR and FNMR on all models, providing an experimental setup for explainable face matching methods.}
\label{tab:patch}
\vspace{-5mm}
\end{table}
\vspace{-3mm}

\vspace{-2mm}
\section{Results}
\vspace{-2mm}
\label{results}
In this section, we present the qualitative and quantitative results, as well as the efficiency of our approach compared to the state-of-the-art. We start with a short qualitative analysis to investigate the visualized explanation maps produced by our approach, xSSAB. We then investigate the performance of our approach and state-of-the-art on the novel Patch-LFW benchmark following the protocol introduced in Section \ref{evalprot}. Last, we present data on the efficiency of our approach and state-of-the-art.
\vspace{-1mm}
\subsection{Qualitative Analysis}
\vspace{-1mm}
To perform the qualitative analysis, we show two example images in Figure \ref{fig:qual}. More examples are presented in the supplementary material. xSSAB explanation map is more fine-grained than the explanation maps from xFace \cite{xface} and xCos \cite{xCos}. In the FNM pair, our approach correctly identified the inserted patch at the nose as not similar but detected the insert front head as similar, which is visually understandable. The explanation maps based on the xFace methods are pretty similar and identified the mouth region as similar, and also the inserted nose patch as dissimilar. Both xCos maps also identified the mouth region as similar. 

For the FM pair, the explanation maps also look very different. We can observe the same different style as in the FNM pair. All the approaches, besides $xCos_{S}$ detect the inserted area in the left mouth region as similar. Our approach and also the xFace methods detect the unchanged area between the eyes as being dissimilar. All previous works limited their evaluation to this visual presentation, however, we believe that such solutions should be evaluated in a more statistically significant quantitative manner as we do in the next section. 

\begin{figure*}
    \centering
    \includegraphics[width=0.50\textwidth]{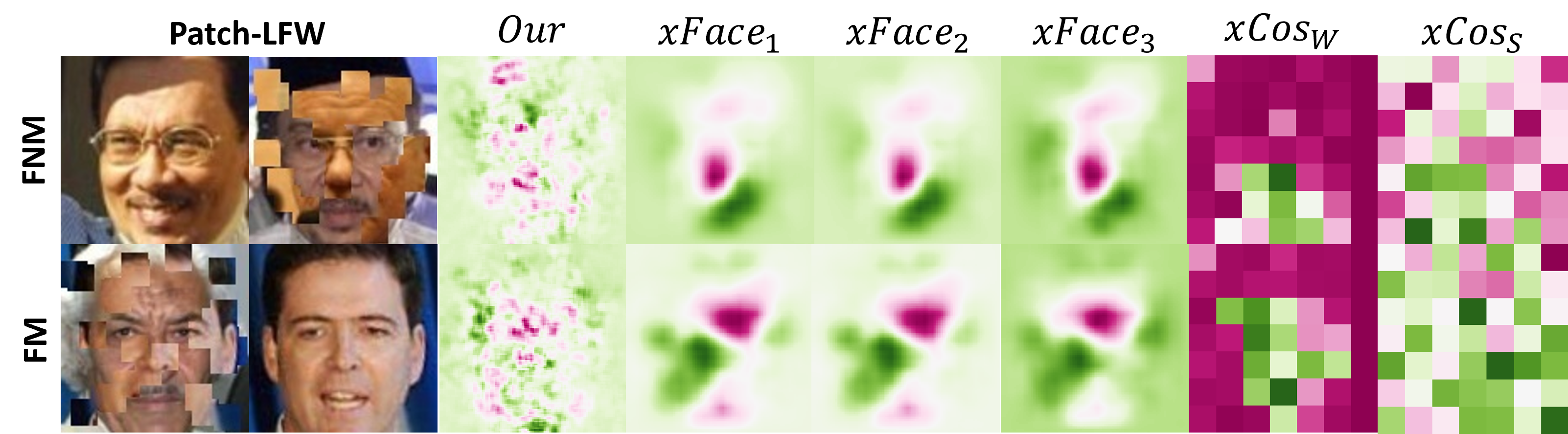}
    \vspace{-2mm}
    \caption{Visualization of the explanation maps of an FNM and an FM pair of Patch-LFW. Our example and the xFace explanation maps are based on the Elasticface-Arc model, the xCos maps are created using the pre-trained module on Arcface. Our approach can detect the inserted patch at the nose as dissimilar to the reference image in the FNM pair. Furthermore, it identified the similar-looking but inserted forehead as similar. In the explanation for the FM, the unchanged area around the eyes is correctly identified as dissimilar.}
    \label{fig:qual} 
    \vspace{-3mm}
\end{figure*}

\begin{figure*}
    \centering
    \includegraphics[width=0.60\textwidth]{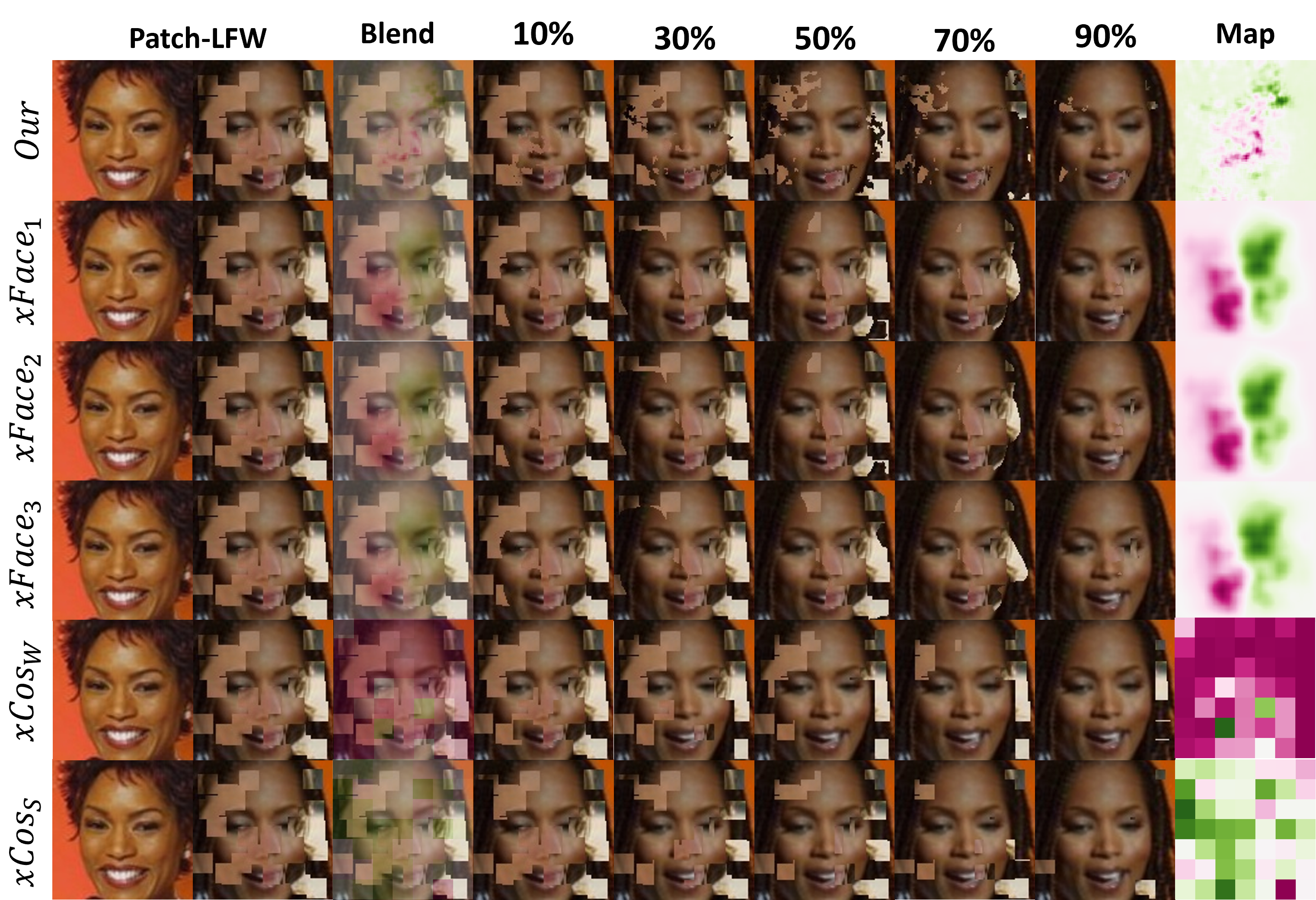}
    \vspace{-2mm}
    \caption{Visualization of the Replacement: Depending on the similarity or dissimilarity, the most influential pixels are replaced with the original pixels from the LFW dataset. Our example and the xFace explanation maps are based on the Elasticface-Arc model, the xCos maps are created using the pre-trained module on Arcface. We can observe, that different areas of the face have been identified as important by the different approaches. Our approach correctly detected the inserted patches at the lower left face side, shortly above the mouth.} 
    \label{fig:replacement}
    \vspace{-4mm}
\end{figure*}

\vspace{-1mm}
\subsection{Quantitative Analysis - DPR Curve}
\vspace{-1mm}
For a quantitative analysis of the quality of our proposed approach and also to compare it objectively with state-of-the-art, we utilize our proposed Patch-LFW and perform the decision-based patch replacement as described in Section \ref{evalprot}. The results for the different approaches on the four utilized models for both error rates, FMR and FNMR are shown in Figure \ref{fig:dpr}{\color{black}, and the AUC values are provided in Table \ref{auc}}.

\begin{table}[h]
\centering
\resizebox{\linewidth}{!}{%
\begin{tabular}{c|cc|cc|cc|cc}
    & \multicolumn{2}{c}{ElasticFace-Arc} & \multicolumn{2}{c}{ElasticFace-Cos} & \multicolumn{2}{c}{CurricularFace} & \multicolumn{2}{c}{ArcFace} \\ \hline
 AUC      & FMR              & FNMR             & FMR               & FNMR            & FMR              & FNMR            & FMR          & FNMR         \\ \hline
$xFace_{1}$ & 1.76             & 5.12             & 2.96              & 2.96            & 3.49             & 2.84            & 4.61         & 7.25         \\
$xFace_{2}$ & 1.67             & 5.06             & 2.88              & 2.94            & 3.51             & 2.85            & 4.58         & 7.83         \\
$xFace_{3}$& 1.63             & 4.81             & 2.92              & 2.87            & 3.62             & 2.80            & 6.86         & 5.31         \\
$xCos_{W}$  & -                & -                & -                 & -               & -                & -               & 9.27         & 11.60        \\
$xCos_{S}$ & -                & -                & -                 & -               & -                & -               & 15.28        & 18.74        \\
Ours   & 2.57             & 8.04             & 5.26              & 5.34            & 5.85             & 4.74            & 8.34         & 12.64        \\
$RND$    & 4.06             & 11.99            & 8.203             & 7.37            & 9.25             & 6.73            & 12.96        & 17.01       \\ \hline
\end{tabular}}
\caption{{\color{black}The AUC values for the curves are shown in Figure \ref{fig:dpr}. A lower AUC indicates better explainability performance. Our proposed approach beats the random approach and the $xCos_{S}$ approach on all models in terms of FMR and FNMR. On the ArcFace model, our training-free approach is competitive with the $xCos_{W}$ approach, which has to be trained in advance. The $xFace$ approaches perform better than our approach but are not efficient, as evaluated and discussed in Section \ref{latency}}}
\label{auc}
\vspace{-6mm}
\end{table}

The results show, that on all models and for both errors, our proposed xSSAB approach outperforms the random baseline {\color{black}(e.g. AUC-FMR of 2.57 to AUC-FMR of 4.06 on random)}. On the ArcFace model, our training-free efficient approach is competitive to the pre-trained $xCos_{W}$ approach {\color{black}(FMR-AUC of 8.34 to 9.27 and FNMR-AUC of 12.64 to 11.60)} and outperforms the pre-trained $xCos_{S}$ approach {\color{black}}. The three black-box based $xFace$ methods outperform our solution, but they require excessive computing power as we will demonstrate later. For the different versions of $xFace$, no clear performance difference can be observed in our quantitative analysis {\color{black}(e.g. ElasticFace-Arc FMR-AUC of 1.76,1.67 and 1.63)}. Our efficient approach shows good performance independent from the utilized underlying FR model. Furthermore, it can be observed, that the detection of the inserted patches on the genuine pairs is easier for the applied methods, as the performance regarding FMR, in general, is better than the performance regarding FNMR. {\color{black}This can be also observed in the AUC values, which are generally higher.}

To the best of our knowledge, with this evaluation, we provided the first quantitative analysis of explainable face verification methods. Using the newly proposed Patch-LFW with the decision-based patch replacement allows us to evaluate the quality of the similarity and dissimilarity maps in a quantified way, even though Patch-LFW is artificially created. An explainable face verification approach, that is capable to correctly explain non-artificial triggers of verification decisions, should also correctly identify artificially added clues.

\subsection{Efficiency}
\vspace{-2mm}
To evaluate the efficiency, we observed the time needed to process and create a single explanation map of a single pair and also to process 100 pairs in total and create the corresponding explanation maps. The times are reported in Table \ref{tab:latency}. We can observe, that the $xFace_{x}$ methods are slow, as they treat the FR model as a black-box and need to manipulate the input multiple times to create meaningful outcomes. In total, 1,130 forward passes are made per image using the proposed parameters by the authors \cite{xface}, leading to 2,260 forward passes for a single explanation map, which is time-consuming at not very efficient (around 12s per single map). The $xCos$ approach is more efficient in terms of creating explanation maps (0.11s per explanation map) but has to be trained in advance. Our training-free approach is much faster than $xFace_{x}$ (only 0.24s per explanation map) as it only requires one forward pass and one backward pass per image. It also outperforms $xCos$ as demonstrated earlier in this section.

\begin{table}[]
\centering
\footnotesize
\begin{tabular}{c|ll}
 Approach & \multicolumn{1}{c}{$T_{Mean}$}  & \multicolumn{1}{c}{$T_{Total}$}  \\ \hline
$xFace_{1}$ & 12.49s            & 1249.69s \\
$xFace_{2}$ & 12.53s            & 1253.31s \\
$xFace_{3}$ & 12.38s            & 1238.87s \\
$xCos$      & 0.11s             & 11.18s  \\
Ours        & 0.24s             & 24.30s  \\ \hline     
\end{tabular}
\vspace{-2mm}
\caption{Evaluation in terms of latency: The average time needed to create an explanation map, as well as the total time needed to create 100 different ones. Our approach is much faster than the other training-free approaches, $xFace$. $xCos$ is faster but has to be trained in advance, which is not represented here and performs worse.} 
\label{tab:latency}
\vspace{-5mm}
\end{table}

\section{Conclusion} 
\vspace{-3mm}
\label{conclusion}
In this work, we proposed xSSAB, an explainable face verification solution based on the backpropagation of similarity score arguments. In our approach, the positive and negative impact of the features on the similarity score based on the FR systems' decision is utilized to obtain visualization of the impact of small, fine-grained face regions on the final matching or non-matching decisions. Our approach efficiently produces fine-grained explanation maps that highlight similar and dissimilar areas as we showed in our experiments and in comparison with two state-of-the-art approaches. To quantitatively evaluate our and other approaches in the field of explainable face verification systems for the first time, we also introduced Patch-LFW. Patch-LFW is a benchmark that is based on LFW \cite{lfw} and has been artificially manipulated to provide more {\color{black}FMs and FNMs} by adding patch-wise clues from another or the same identity. An explainable face verification system can now be evaluated based on the Decision-based Patch Replacement (DPR) curve, which represents how good an explainability solution does its task. In an evaluation of latency, we also showed the efficiency of our approach, which is also training-free in comparison to black-box approaches, which are time-consuming by design. As FR systems become more ubiquitous in our daily lives and the precise workings of highly complex deep-learning-based models remain difficult to comprehend, increasing explainability, interpretability, and transparency will continue to be important, especially in the biometric area, as personal and private data is processed.
\vspace{-5mm}
\paragraph{Acknowledgement}
This research work has been funded by the German Federal Ministry of Education and Research and the Hessian Ministry of Higher Education, Research, Science, and the Arts within their joint support of the National Research Center for Applied Cybersecurity ATHENE. This work has been partially funded by the German Federal Ministry of Education and Research through the Software Campus Project.

{\small
\bibliographystyle{ieee_fullname}
\bibliography{egbib}
}

\end{document}